\documentclass[letterpaper, 10 pt, conference]{ieeeconf}  

\IEEEoverridecommandlockouts                              

\usepackage[T1]{fontenc}
\usepackage{xcolor}
\usepackage{hyperref}
\usepackage{cite}
\usepackage{graphicx}
\usepackage{graphbox}
\usepackage{subcaption}
\usepackage{textcomp}
\usepackage{amsmath}
\usepackage{multirow}
\begin{document}

\title{\LARGE \bf
TGRMPT: A Head-Shoulder Aided Multi-Person Tracker and a New Large-Scale Dataset for Tour-Guide Robot
}

\author{Wen Wang, Shunda Hu, Shiqiang Zhu, Wei Song, Zheyuan Lin, Tianlei Jin, Zonghao Mu, Yuanhai Zhou
	\thanks{All the authors are with the Intelligent Robotics Research Center, Zhejiang Lab, China.}
	\thanks{E-mail: [wangwen, hushunda, zhusq, weisong, linzhy, jtl, muzonghao, zhouyh]@zhejianglab.com}
}

\maketitle
\thispagestyle{empty}
\pagestyle{empty}

\begin{abstract}

A service robot serving safely and politely needs to track the surrounding people robustly, especially for Tour-Guide Robot (TGR). However, existing multi-object tracking (MOT) or multi-person tracking (MPT) methods are not applicable to TGR for the following reasons: 1. lacking relevant large-scale datasets; 2. lacking applicable metrics to evaluate trackers. In this work, we target the visual perceptual tasks for TGR and present the TGRDB dataset, a novel large-scale multi-person tracking dataset containing roughly 5.6 hours of annotated videos and over 450 long-term trajectories. Besides, we propose a more applicable metric to evaluate trackers using our dataset. As part of our work, we present TGRMPT, a novel MPT system that incorporates information from head shoulder and whole body, and achieves state-of-the-art performance. We have released our codes and dataset in \href{https://github.com/wenwenzju/TGRMPT}{https://github.com/wenwenzju/TGRMPT}.

\end{abstract}

\section{INTRODUCTION}

Service robots work in human-populated environments. Robust tracking surrounding people allow robots to serve and navigate safely and politely. In this work, we aim at the egocentric perceptual task, multi-person tracking, using a tour-guide robot (TGR). The TGR is a robot that can show people around a site such as museums, castles, and aquariums, and can introduce surroundings to people \cite{al2016tour}. In this scenario, targets occlude with each other and step in or out the camera view, frequently. Moreover, humans in a group usually dress in the same clothes, making it a difficult task to identify each other. Despite the above challenges, individuals in a guided group must be tracked and assigned consistent IDs during the whole tour-guide service period.

However, there is a vast gap between the existing MOT or MPT datasets and the tour-guide scenario, making existing trackers perform unsatisfactorily.	Some examples of such datasets are the well-known MOTChallenge \cite{milan2016mot16} and KITTI \cite{geiger2012we}, targeting applications in video surveillance and autonomous driving, respectively. Furthermore, in these datasets, a person will be assigned a new ID if he/she leaves the field of view or is occluded for a prolonged period and then reappears. This is very different from our scenario where consistent IDs are required during the whole tracking period. More recently, an egocentric dataset, JRDB, was presented in \cite{martin2021jrdb}, which is more relevant to our work. JRDB is a multi-modal dataset and was captured using a moving robot in a university campus, both indoors and outdoors. However, this dataset is still inapplicable in our scenario. On the one hand, JRDB was captured in daily life where few interactions between students and robot exist. As a result, cases that humans disappear then reappear again are few. On the other hand, people in JRDB dress casually and almost no one wears the same.

\begin{figure}[!t]
	\centering
	\includegraphics[width=0.45\textwidth]{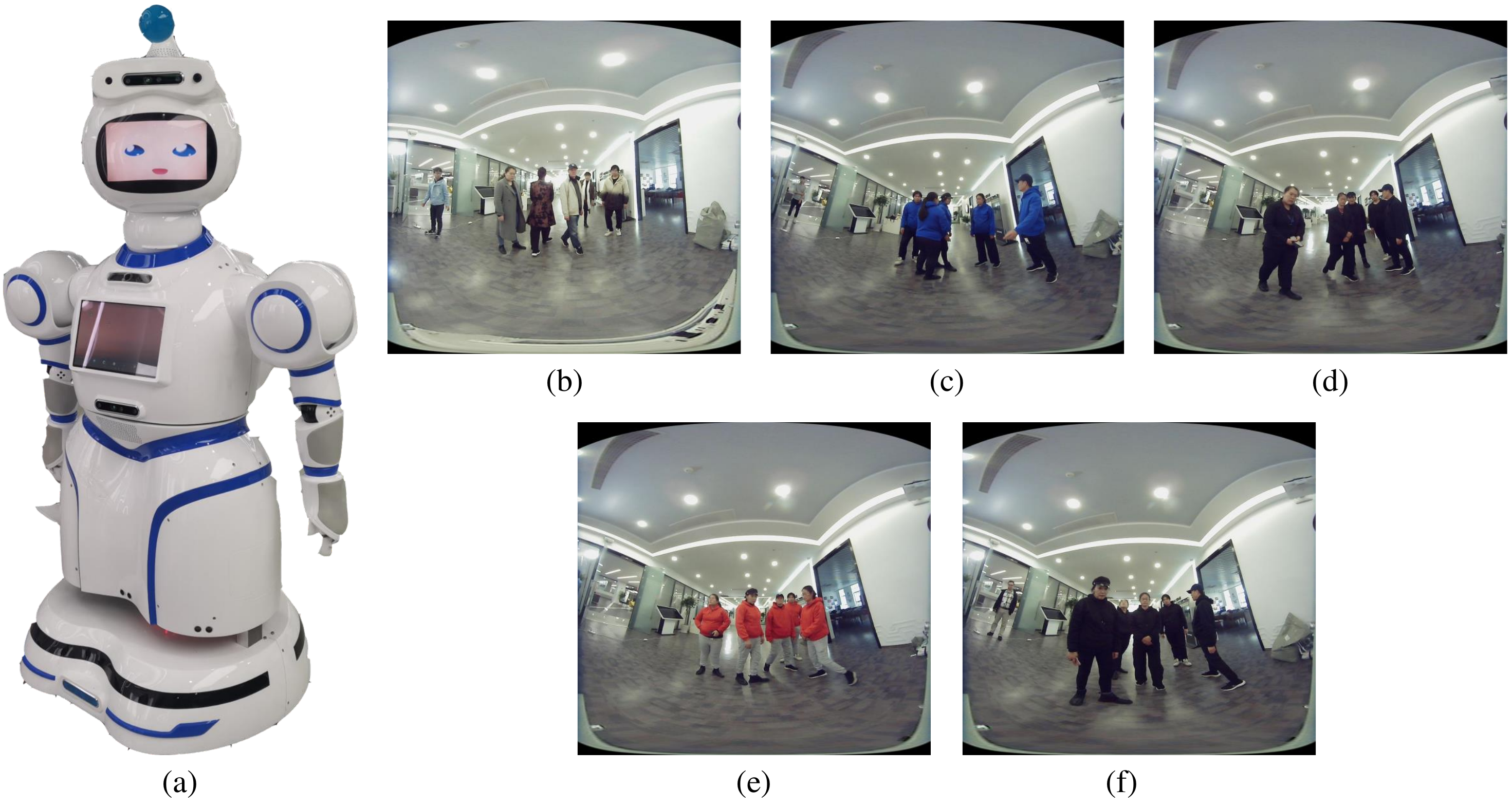}
	\caption{(a) The tour-guide robot that we use in this paper. (b)-(f) Examples of our dataset captured in 5 rounds. In the first round, people wear their own clothes. In the following 4 rounds, people wear same clothes.}.
	\label{intro}
\end{figure}

To solve the above problems, we present the TGRDB dataset, a new large-scale dataset for TGR. To capture our dataset, we use a $180^{\circ}$ fisheye RGB camera mounted on a moving (sometimes standing) robot shown in Fig. \ref{intro}. In an indoor tour-guide scenario, the robot shows 5 or 6 individuals around. At first, participants wear their own clothes and complete the first capturing round. Then they are required to change and put on the same clothes, as shown in Fig. \ref{intro}. There are totally 5 capturing rounds for each group, including one casual-cloth round and 4 same-cloth rounds, and there are 18 groups in total. As a result, 90 video sequences are captured, totally of 5.6 hours long. We annotate each video frame with whole body and head shoulder bounding boxes, and unique IDs, resulting in over 3.2 million bounding boxes and 450 trajectories. We hope this dataset will drive the progress of research in service robotics, long-term multi-person tracking, and fine-grained \cite{yin2020fine} or clothes-inconsistency \cite{wan2020person} person re-identification. Associated with this dataset, a new metric named TGRHOTA is proposed. It is a simplified but more practical version of HOTA \cite{luiten2021hota}. TGRHOTA only considers matches between prediction and ground truth as true positives at the most previous frames when evaluating association score. We will describe it in Sec. \ref{metric}.

As part of our work, we present TGRMPT, a novel and practical multi-person tracker in the tour-guide scenario. As a person's head-shoulder contains discriminative information such as hairstyle \cite{xu2020black}, and involves fewer distractors than the whole body bounding box when he/she is partially occluded, we optimally integrate head-shoulder information into the tracker. Specifically, we adopt the state-of-the-art DeepSORT \cite{wojke2017simple} framework and tailor it to our scenario. Both whole body and head shoulder detectors are trained using our dataset, as well as deep appearance embedding feature extractors. To integrate these two types of information, we simply concatenate the two embedding features extracted from the corresponding bounding boxes, resulting in stronger appearance descriptors. Extensive experiments verify the effectiveness of our proposal.

To summarize, our contributions are as follows:
\begin{itemize}
	\item We release the TGRDB dataset, a first large-scale dataset towards the applications of the tour-guide robot. This dataset not only benefits the domain of service robotics but can also drive the progress of domains related to multi-person tracking and person re-identification.
	\item We propose a more practical metric, TGRHOTA, to evaluate trackers in the tour-guide scenario. Different from existing metrics, TGRHOTA punishes trackers when a new ID assignment occurs if the target has already been assigned an ID before.
	\item We propose a novel head-shoulder aided multi-person tracker, named TGRMPT, that leverages best of both information containing in the whole body and head shoulder. Experiments show that our tracker achieves start-of-the-art results.
\end{itemize}

\section{Related Work}

\subsection{Tracking Methods}

Most multi-object tracking algorithms consist of two critical components, object detection and data association, responsible for estimating the bounding boxes and obtaining the identities, respectively.

\textbf{Detection in tracking}: Benefiting from the development of deep learning, the detection algorithm has been significantly improved. In addition, the quality of detection directly affects the tracking results. Therefore, many researchers \cite{yu2016poi,xu2019spatial,zhou2020tracking,sun2020transtrack,xu2021transcenter} are committed to improving the detection ability. 
POI \cite{yu2016poi} achieves the state-of-the-art tracking performance due to the high-performance detection and deep learning-based appearance features.
STRN \cite{xu2019spatial} presents a similarity learning framework between tracks and objects, which encodes various Spatial-Temporal relations. The tracking-by-detection pipeline achieves leading performance, but the model complexity and computational cost are not satisfying.
Centertrack \cite{zhou2020tracking} proposes to estimate detection box and offset by using the data of two adjacent frames, which improves the ability to recover missing or occluded objects.
\cite{sun2020transtrack} leverages the transformer architecture, an attention-based query-key mechanism, to integrate the detection information of the previous frame to the detection process of the current frame.
\cite{xu2021transcenter} leverages two adjacent frames information to improve detection under the transform framework and proposes dense pixel-level multi-scale queries that are mutually correlated within the transformer attention and produce abundant but less noisy tracks.                                                         
In addition, many works directly use the off-shelf detection methods, e.g., two-stage \cite{ren2015faster} or one-stage object detectors \cite{lin2017focal}, YOLO series detectors \cite{redmon2018yolov3}, anchor-free detectors \cite{law2018cornernet}.

\textbf{Data association}: As the core component of multi-object tracking, data association first computes the similarity between tracklets and detection boxes, then matches them according to the similarity. Many methods focus on data association to improve tracking performance. SORT~\cite{bewley2016simple} is a simple but effective tracking framework that employs Kalman filtering in image space and frame-by-frame data association using the Hungarian method with the association metric that measures bounding box overlap. To address occlusions, \cite{wojke2017simple} put forward to adopt an independent REID model to extract appearance features from the detection boxes to enhance the association metric of SORT. 
To save computational time, \cite{zhang2021fairmot,wang2020towards,liang2020rethinking,lu2020retinatrack} integrate the detecting and embedding models into a single network.
To address the non-negligible true object missing and fragmented trajectories that are caused by simply throwing away the objects with low detection scores, ByteTrack \cite{zhang2021bytetrack} propose to track by associating all detection boxes. To recover true objects, the association method in \cite{zhang2021bytetrack} utilizes similarities to filter out background detections with low scores. 
Many researchers consider splitting the whole tracking task into isolated sub-tasks, such as object detection, feature extraction, and data association, which may lead to local optima. To address this issue, the research in \cite{peng2020chained}, as well as the subsequent works in \cite{zhou2020tracking,tokmakov2021learning,pang2021quasi}, propose to use an end-to-end model to unify the three isolated subtasks.

\subsection{Tracking Datasets}
Our TGRDB is a multi-person tracking dataset and benchmark containing fine-grained and clothes-inconsistency targets in the tour-guide scenario. \cite{milan2016mot16,dendorfer2020mot20,martin2021jrdb,yang2019person,yin2020fine} are the most relevant datasets to TGRDB. MOT \cite{milan2016mot16,dendorfer2020mot20} is a well-known multi-object tracking benchmark, and many methods are based on this. JRDB \cite{martin2021jrdb} is a novel multi-modal dataset collected from a mobile social JackRabbot. \cite{yang2019person} is an image dataset for cloth-changing person REID. \cite{yin2020fine} released a fine-grained REID dataset containing targets with the same clothes. In addition, the automatic driving datasets\cite{geiger2012we,yu2018bdd100k,sun2020scalability} are also interrelated to our TGRDB, which annotate a large number of pedestrians bounding boxes. To the best of our knowledge, there is no tracking dataset containing fine-grained and clothes-inconsistency targets on the same scale as our TGRDB in the tour-guide scenario. 

\subsection{Metrics}
Within the last few years, new MOT metrics have been proposed enormously.
To remedy the lack of generally applicable metrics in MOT, \cite{bernardin2008evaluating} introduces two novel metrics, the multi-object tracking precision (MOTP) and the multi-object tracking accuracy (MOTA), that intuitively express a tracker’s overall strengths.
\cite{ristani2016performance} proposes a new precision-recall measure of performance, IDF1, that treats errors of all types uniformly and emphasizes correct identification over sources of errors.
MOTA and IDF1 overemphasize the importance of detection and association separately. To address this, \cite{luiten2021hota} presents a novel MOT evaluation metric, higher-order tracking accuracy (HOTA), which is a unified metric that explicitly balances the effect of detection, association, and localization. Nonetheless, none of the above metrics is applicable in our scenario. Hence we propose a new metric, TGRHOTA, for fair comparison of trackers in TGR.

\section{Dataset and Metric}

\subsection{Dataset}

Our TGRDB dataset was collected with a $180^\circ$ fisheye RGB camera on-board of a tour-guide robot shown in Fig. \ref{intro}. Equipped with a $360^\circ$ 2D LiDAR, our TGR can autonomously navigate and show people around. There is a pan-tilt with two degrees of freedom mounted at the neck, so the robot can rotate its head to see what it's interested in.

\begin{figure}[th]
	\centering
	\includegraphics[width=0.45\textwidth]{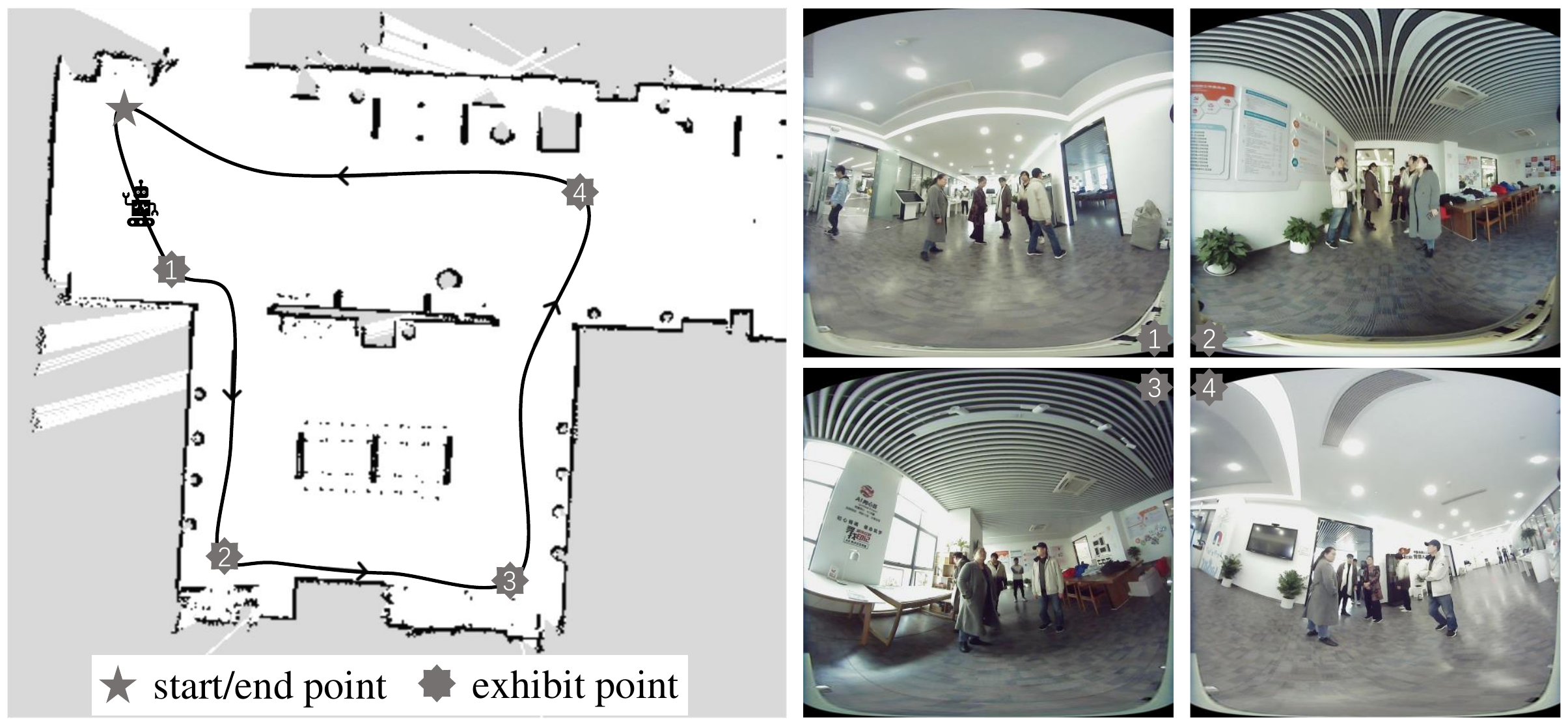}
	\caption{The tour-guide route to capture our TGRDB dataset. There are 4 exhibit points where the robot stays and introduces exhibits to participants. Pictures are captured at these points respectively.}.
	\label{route}
\end{figure}

We collect our dataset in an indoor tour-guide scenario. From the start point, the robot navigates to the first preset exhibit point, stays for a while and introduces the surrounding exhibits to participants, then moves to the next exhibit point, as shown in Fig. \ref{route}. Repeat the above steps until it returns to the start point. There are totally four exhibit points and the robot stays for around 22 seconds at each point. During the introduction period, the head of the robot alternatively looks forward or rotates randomly. Participants are required to walk freely and behave naturally. They need to follow the robot when it moves from one exhibit point to another. During this period targets are out of the field of camera view. 

We divide all the participants into 18 groups, in each there are 5 or 6 individuals. Each group is required to finish 5 tour-guide rounds, including 1 casual-cloth round and 4 same-cloth rounds as shown in Fig. \ref{intro}.

Each video frame is annotated with both whole-body and head-shoulder bounding boxes, as well as unique IDs. For one target, his/her ID is consistent no matter which cloth he/she dresses in. As a result, totally 90 video sequences are captured at 25 fps. The average duration of each video is 3.76 minutes. We divide the dataset into train and test sub-datasets. Table \ref{dataset} shows more details of our dataset.

\begin{table}[ht]
	\centering
	\caption{Statistical comparisons between TGRDB and existing datasets. K = Thousand, M = Million, min = minutes.}
	\label{dataset}
	\setlength{\tabcolsep}{0.3mm}{
	\begin{tabular}{l|c|c|c|c|c|c}
		\hline
		& No. of & No. of & No. of & No. of & & \\ 
		& sequences & frames & boxes & IDs & Duration & Cloth \\ \hline
		MOT17\cite{milan2016mot16} & 14 & 34K & 290K & - & - & Casual only \\ 
		MOT20\cite{dendorfer2020mot20} & 8 & 13K & 1.6M & - & - & Casual only \\ \hline
		KITTI\cite{geiger2012we} & 22 & 15K & 6K & - & - & Casual only \\
		BDD100K\cite{yu2018bdd100k} & 1.6K & 318K & 440K & - & - & Casual only \\
		Waymo\cite{sun2020scalability} & 1.15K & 600K & 2.1M & - & - & Casual only \\ \hline
		JRDB\cite{martin2021jrdb} & 54 & 28K & 2.4M & - & 64min & Casual only \\
		Ours(Train) & 50 & 281K & 1.8M & 51 & 188min & Casual \& Same \\ 
		Ours(Test) & 40 & 232K & 1.4M & 40 & 150min & Casual \& Same \\
		\textbf{Ours(Total)} & \textbf{90} & \textbf{513K} & \textbf{3.2M} & \textbf{91} & \textbf{338min} & \textbf{Casual \& Same} \\
		\hline
	\end{tabular}}
\end{table}

\subsection{Metric} \label{metric}

\begin{figure}[!t]
	\centering
	\includegraphics[width=0.45\textwidth]{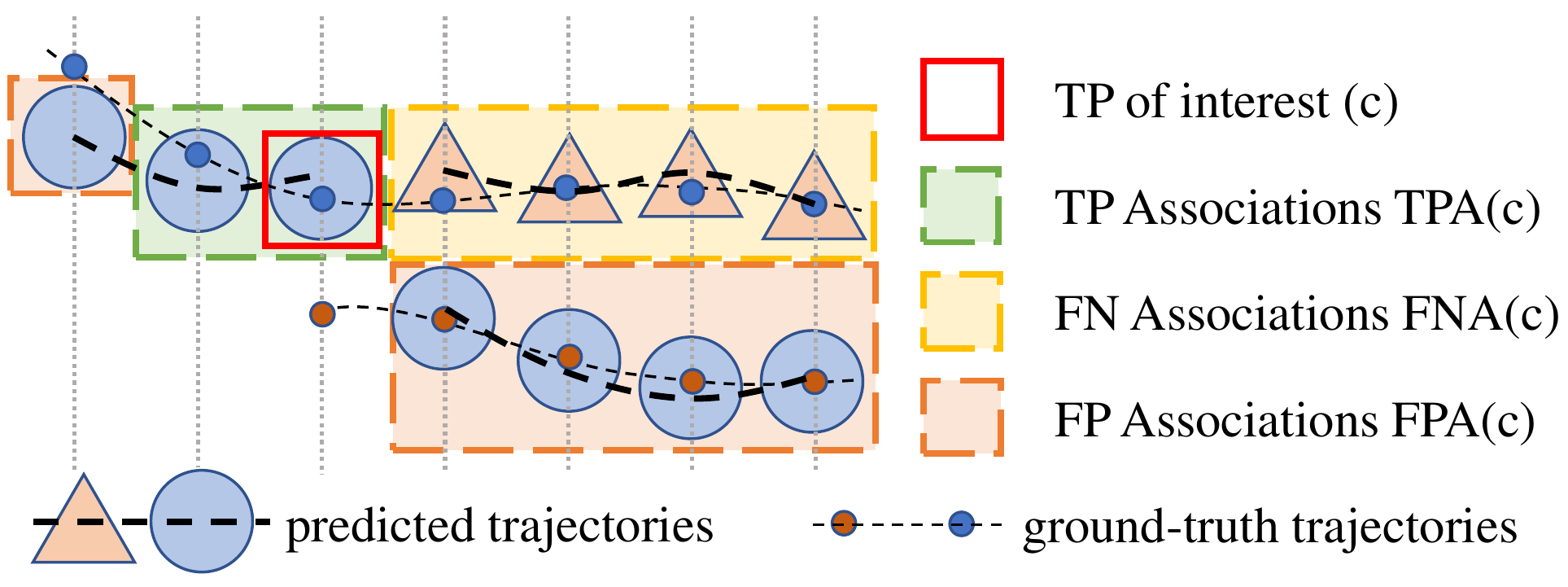}
	\caption{Unlike existing metrics, in TGRHOTA, one predicted trajectory can be matched to only one ground-truth trajectory, and vice versa. In other words, existing metrics select TP of interest in all dashed areas, while our TGRHOTA only selects in the green area.}.
	\label{metric_fig}
\end{figure}

\begin{figure}[t]
	\centering
	\includegraphics[width=0.45\textwidth]{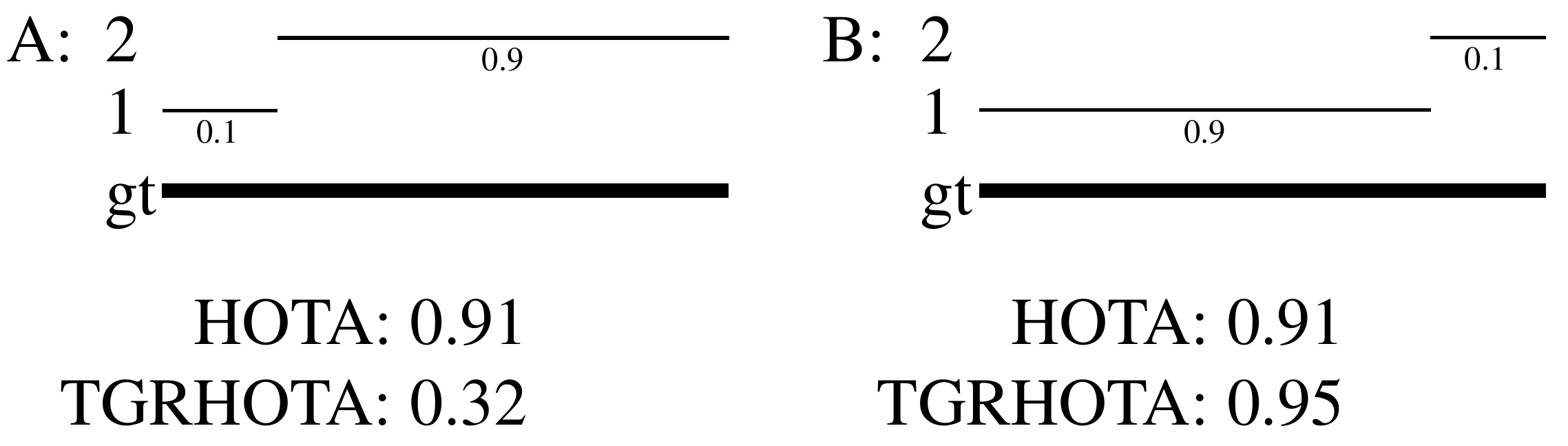}
	\caption{An example showing how HOTA fails to correctly rank tracking performance in TGR because it takes all TPs into account. Thick line: ground-truth trajectory. Thin lines: prediction trajectories. All detections are TPs.}
	\label{hota_vs_tgrhota}
\end{figure}

An applicable metric is significant when comparing different trackers. Existing metrics, such as HOTA  \cite{luiten2021hota}, MOTA \cite{bernardin2008evaluating} and IDF1 \cite{ristani2016performance}, treat all matched prediction (pr) and ground-truth (gt) pairs as true positives (TPs), which may be reasonable in video surveillance or autonomous driving. However, they are not applicable in TGR for the requirement of consistent IDs during the whole tour-guide period. In retrospect, HOTA \cite{luiten2021hota} counts TPAs (True Positive Associations), FNAs (False Negative Associations), and FPAs (False Positive Associations) for each TP and defines association score as follows: 
\begin{equation}
	AssA=\frac{1}{|TP|}\sum_{c\in TP}{\frac{|TPA(c)|}{|TPA(c)|+|FNA(c)|+|FPA(c)|}}, 
\end{equation}
where all matched pr and gt pairs are considered as TP of interest. In other words, as shown in Fig. \ref{metric_fig}, TP of interest is selected in matches within all dashed areas. However, in TGR, we only consider TPs in the green area, i.e., matches at most previous frames. Formally, we define the set of TPs in TGR as follows:
\begin{equation}
	TP'=\{tp_t\in TP|\forall t'<t, tp_{t'}\equiv tp_t \text{ or } tp_{t'}\not\equiv tp_t\}
\end{equation}
where $t$ is the time index, and $\equiv$ denotes that two TPs have the same prID and gtID, while $\not\equiv$ denotes two TPs have different prID and gtID. Thus, the association score in TGR can be written as:
\begin{equation}
	AssA'=\frac{1}{|TP'|}\sum_{c\in TP'}{\frac{|TPA(c)|}{|TPA(c)|+|FNA(c)|+|FPA(c)|}}.
\end{equation}

The detection score, $DetA$, and the calculation of the final HOTA score, is the same as in \cite{luiten2021hota}, except that we use $AssA'$ instead of $AssA$.

Fig. \ref{hota_vs_tgrhota} shows an example where HOTA gives the two trackers the same scores which is counterproductive for our TGR scenario where tracker B performs much better than tracker A. Treating all TPs equally even if they have been assigned another ID at a previous frame, HOTA is not applicable in our scenario. On the contrary, our TGRHOTA gives much more reasonable scores.

\section{TGRMPT: Head-Shoulder Aided Multi-Person Tracking}

\begin{figure*}[t]
	\centering
	\includegraphics[width=0.95\textwidth]{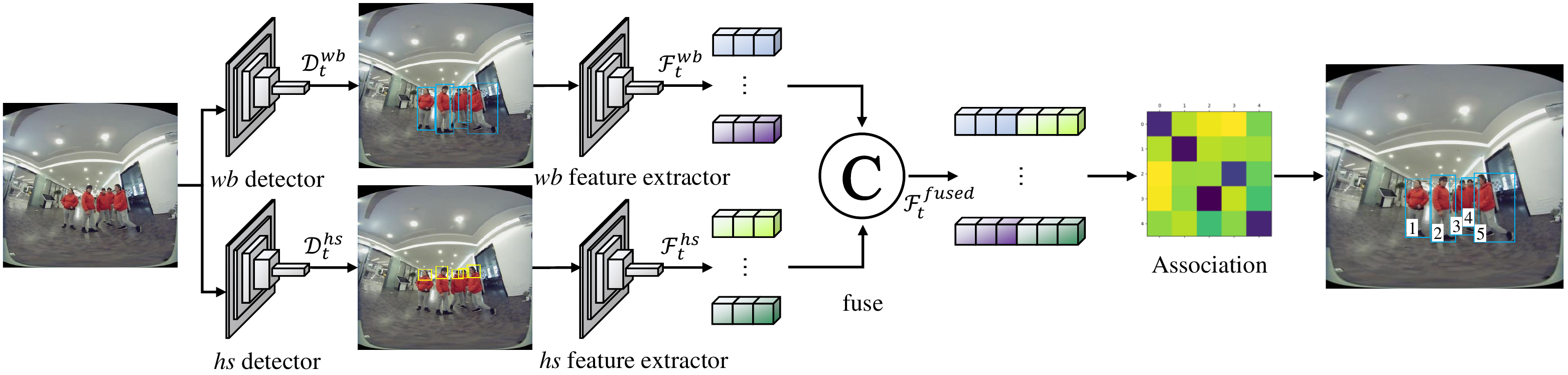}
	\caption{The pipeline of our proposed TGRMPT framework. Integrating both whole body (\textit{wb}) and head shoulder (\textit{hs}) information, our method produces strong and discriminative appearance feature descriptors. Followed by Hungarian algorithm, robust tracking result is obtained.}
	\label{pipeline}
\end{figure*}

Fig. \ref{pipeline} depicts our proposed TGRMPT tracker. By sequentially applied whole body (\textit{wb})/head shoulder (\textit{hs}) detectors and \textit{wb}/\textit{hs} feature extractors, deep appearance signatures describing global whole-body and local head-shoulder are generated. Concatenating these two types of features results in the final strong descriptor that contains both global and local information. The followed Hungarian algorithm produces the final robust tracking result.

\subsection{Detector}

Detection is the core component of the existing MOT or MPT systems. In consider of system speed, we deploy YOLOv5s \cite{glenn_jocher_2021_5563715} as our detector. We fine-tune this network on our dataset using \textit{wb} and \textit{hs} annotations to form \textit{wb} and \textit{hs} detectors, respectively. The corresponding outputs are denoted as $\mathcal{D}_t^{wb}=\{d_1^{wb},...,d_N^{wb}\}$ and $\mathcal{D}_t^{hs}=\{d_1^{hs},...,d_M^{hs}\}$. Defining IoU between $d_i^{wb}$ and $d_j^{hs}$ as $IoU=\frac{|d_i^{wb}\cap d_j^{hs}|}{|d_j^{hs}|}$, we use Hungarian algorithm to match \textit{wb} and \textit{hs} detections, resulting in matched pairs denoted as $\mathcal{D}_t^{match}=\{(d_{i1}^{wb},d_{j1}^{hs}),...,(d_{iK}^{wb},d_{jK}^{hs})\}$. We denote those \textit{wb} detections that are not matched to any \textit{hs} detections as $\mathcal{D}_t^{wb^-}=\{d_1^{wb^-},...,d_L^{wb^-}\}$, and we discard those \textit{hs} detections that are not matched to any \textit{wb} detections.

\subsection{Appearance Descriptor}

Appearance descriptors are used to measure the similarities between detections at current frame and history trajectories. They offer signification, sometimes the only, cue to re-identify the target when he/she was missed for a period of time and reappears. To obtain strong and discriminative appearance descriptors, we leverage ResNet18 \cite{he2016deep} as our feature extracting network and train it using our TGRDB dataset described above. We choose ResNet18 due to the trade-off between speed and performance. To integrate global and local information, \textit{wb} and \textit{hs} feature extracting networks are trained individually using \textit{wb} and \textit{hs} re-identification dataset, respectively. For detection pair $(d_i^{wb},d_j^{hs})\in \mathcal{D}_t^{match}$, the corresponding appearance features are $(f_i^{wb}, f_j^{hs})$. For detection $d_i^{wb^-}\in\mathcal{D}_t^{wb^-}$, we denote its appearance feature as $(f_i^{wb^-}, \mathbf{0})$ where $\mathbf{0}$ is zero vector with the same dimension as $f_i^{wb^-}$. At last, we concatenate each pair of features and form the final appearance descriptors denoted as $\mathcal{F}_t^{fused}=\{[f_{i1}^{wb},f_{j1}^{hs}],...,[f_{iK}^{wb},f_{jK}^{hs}],[f_1^{wb^-}, \mathbf{0}],...,[f_L^{wb^-}, \mathbf{0}]\}$.

\subsection{Tracking} \label{tracking}

We adopt DeepSORT \cite{wojke2017simple} framework to perform tracking. Given a set of appearance descriptors, $\mathcal{F}_t^{fused}$, we associate them to trajectories at previous frames. To do so, we keep the latest $P$ descriptors for each trajectory and leverage cosine similarity to measure the distance between trajectories and detections. Specifically, for one detection $f\in \mathcal{F}_t^{fused}$ and one trajectory $\mathcal{T}=\{f_1,...,f_P\}$, a set of cosine similarities, $S(f,\mathcal{T})=\{s_1,...,s_P\}$ are calculated, and we average them to produce the final distance, $D(f,\mathcal{T})=\sum_{i=0}^{P}(1-s_i)/P$. We also explore the performance of the minimum similarity values, $D(f,\mathcal{T})=min(1-s_i)$, in experiments. After computing the pairwise distance between current detections and history trajectories, we obtain $Q\times (K+L)$ cost matrix where $Q$ is the number of trajectories, i.e., the number of tracked targets. Hungarian algorithm is employed to get the final association results and trajectories are updated using the corresponding associated detections. Any association that has a distance greater than the preset threshold $\tau$ is treated as unmatched.

The origin DeepSORT is designed for video surveillance and is not well fitted to our tour-guide scenario. In the data association process of DeepSORT, matching cascade is introduced to give higher priority to trajectories that miss targets for less time. Meanwhile, a so-called gating mechanism is applied and only detections near the predicted locations of trajectories are considered. These two tricks may cause more track fragments and worsen the tracking performance in our scenario as frequent occlusion and long-term missing exist in our dataset. Consequently, we abandon these two tricks in our tracker. Another trick to handle challenges in TGR is to assign a relatively great value to the age threshold, $\alpha$. When a trajectory is missed, i.e., no detections associated with it, for consecutive $\alpha$ frames, it will be deleted. We explore different values of $\alpha$ in our ablation experiments and conclude that $\alpha=\infty$, i.e., no trajectories are deleted during tracking, gives the best performance.

\section{Experiments}

To evaluate on our TGRDB dataset, the widely used metrics, HOTA \cite{luiten2021hota}, MOTA \cite{bernardin2008evaluating} and IDF1 \cite{ristani2016performance} are adopted. Besides, we report results evaluated using the proposed TGRHOTA metric as well. For a more comprehensive analysis, we split the test dataset into casual-cloth and same-cloth sub-datasets, and conduct experiments respectively. But we don't split the training dataset.

\subsection{Ablation Studies}

\subsubsection{Hyper-Parameters}

Different values of hyper-parameters, $\alpha$ and $\tau$, may impact the final performance dramatically. We fix the distance threshold $\tau$ and set it to 0.5 at first, and change the age threshold $\alpha$ to explore the impact. In this stage, we employ the whole body branch only. The results are shown in Table \ref{alpha}. Intuitively, $\alpha$ impacts more on data association. We can see this from AssA and IDF1 scores. In the TGR scenario, it is required that people's IDs remain consistent even after long-term target missing. Thus the greater $\alpha$ should give the better performance. This is verified by the results in Table \ref{alpha}, where the infinite $\alpha$ gives the best performance. This means that during the period of tour-guide, no tracks will be deleted, which gives chance to the tracker to find back the target even if he/she has disappeared for a long time.

\begin{table}[th]
	\centering
	\caption{The impact of age threshold $\alpha$ when distance threshold $\tau=0.5$.}
	\label{alpha}
	\begin{subtable}[t]{0.495\linewidth}
		\caption{Casual Cloth}
		\centering
		\setlength{\tabcolsep}{1mm}{
			\begin{tabular}{l|cccc}
				\hline
				& \multicolumn{4}{|c}{$\alpha$} \\
				& 30 & 100 & 500 & $\infty$ \\ \hline
				HOTA$\uparrow$ & 33.6 & 36.0 & 37.8 & \textbf{41.2} \\ \hline
				AssA$\uparrow$ & 15.0 & 17.1 & 18.9 & \textbf{22.5} \\ \hline
				MOTA$\uparrow$ & 86.2 & 86.3 & 86.0 & 85.5 \\ \hline
				IDF1$\uparrow$ & 27.4 & 29.6 & 31.2 & \textbf{36.4} \\ \hline
		\end{tabular}}
	\end{subtable}
	\begin{subtable}[t]{0.495\linewidth}
		\caption{Same Cloth}
		\centering
		\setlength{\tabcolsep}{1mm}{
			\begin{tabular}{l|cccc}
				\hline
				& \multicolumn{4}{|c}{$\alpha$} \\
				& 30 & 100 & 500 & $\infty$ \\ \hline
				HOTA$\uparrow$ & 31.7 & 34.8 & 37.2 & \textbf{42.5} \\ \hline
				AssA$\uparrow$ & 13.1 & 15.8 & 18.1 & \textbf{23.6} \\ \hline
				MOTA$\uparrow$ & 86.9 & 87.0 & 86.9 & 86.4 \\ \hline
				IDF1$\uparrow$ & 24.9 & 28.4 & 30.8 & \textbf{37.7} \\ \hline
		\end{tabular}}
	\end{subtable}
\end{table}

 \begin{figure}[th]
	\centering
	\includegraphics[width=0.45\textwidth]{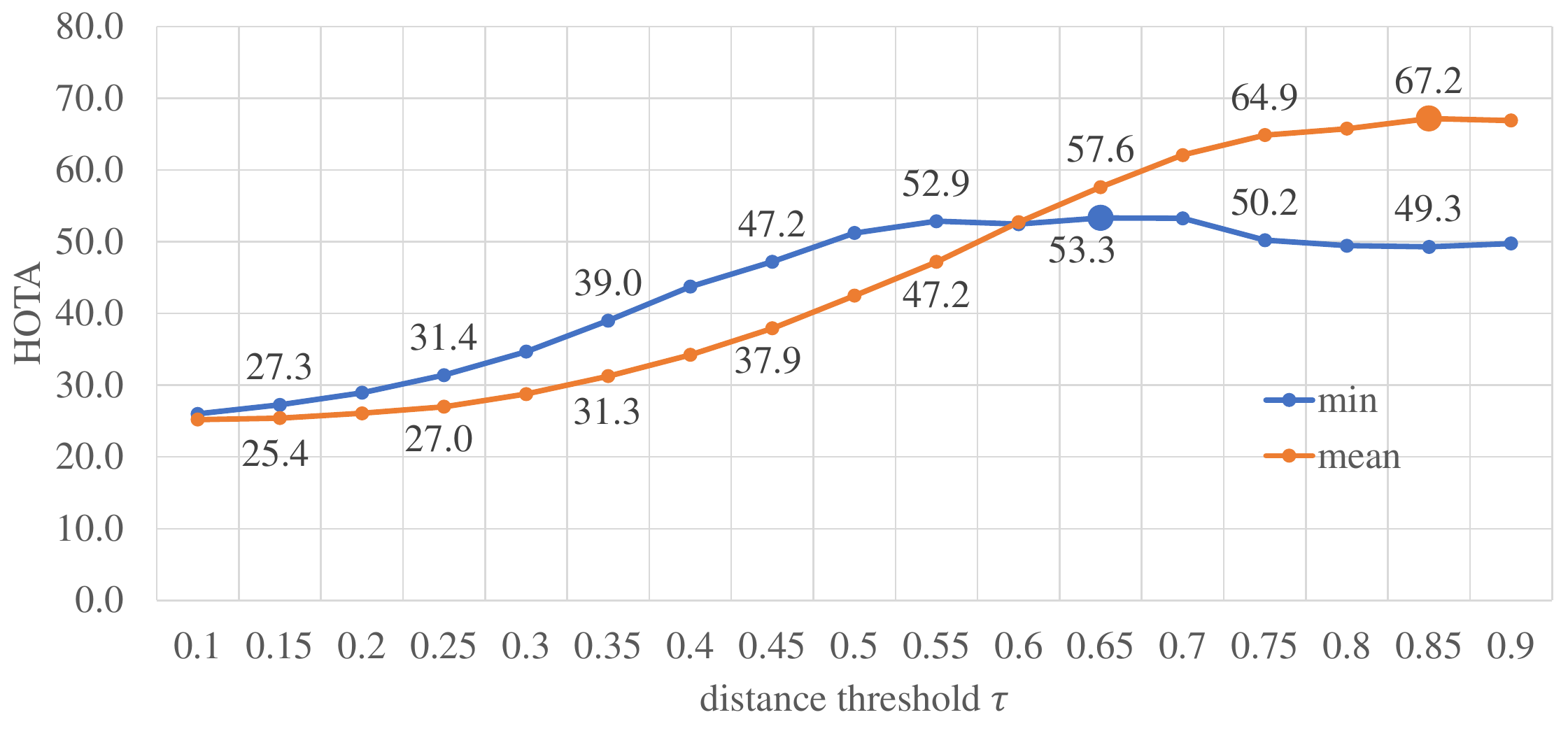}
	\caption{The impact of different distance threshold $\tau$ values using the two matching distance calculation ways.}.
	\label{distance}
\end{figure}

We then fix $\alpha=\infty$ and change the distance threshold $\tau$ to explore the impact. In data association, there are two ways, \textit{mean} or \textit{min} as described in \ref{tracking}, to compute the distance between detection and trajectory. We show their results\footnote{For simplicity, only results on same-cloth sub-dataset are shown as casual-cloth sub-dataset has similar results.} in
Fig. \ref{distance}, where we conclude that the averaged distance value gives much better performance. The reason is apparent: the reserved $P$ galleries in each trajectory may contain noisy samples and the minimum distance value suffers much from these noises.

The above experiments on hyper-parameters indicate that $\alpha=\infty$ and $\tau=0.85$ give the best performance. We will fix them in the following experiments and explore other aspects of our method.

\subsubsection{Addition of Head-Shoulder}

Head-shoulder contains non-negligible features, such as haircut, glasses, complexion, etc., which can provide discriminative cues for person re-identification. We conduct comparative experiments on the employment of: 1) \textit{wb}, the whole body only; 2) \textit{hs}, the head shoulder only; 3) \textit{wb}+\textit{hs}, both whole body and head shoulder. The results are shown in Table \ref{wb_hs}. Undoubtedly, the method that integrates both \textit{wb} and \textit{hs} information gives the best performance. 

\begin{table}[th]
	\centering
	\caption{The improvement of performance after adding head-shoulder.}
	\label{wb_hs}
	\begin{subtable}[t]{0.495\linewidth}
		\caption{Casual Cloth}
		\centering
		\setlength{\tabcolsep}{1.5mm}{
			\begin{tabular}{l|ccc}
				\hline
				& \textit{wb} & \textit{hs} & \textit{wb}+\textit{hs} \\ \hline
				HOTA$\uparrow$ & 71.2 & 53.0 & \textbf{72.7} \\ \hline
				MOTA$\uparrow$ & 87.1 & 54.1 & \textbf{87.2}  \\ \hline
				IDF1$\uparrow$ & 81.2 & 58.8 & \textbf{84.2} \\ \hline
		\end{tabular}}
	\end{subtable}
	\begin{subtable}[t]{0.495\linewidth}
		\caption{Same Cloth}
		\centering
		\setlength{\tabcolsep}{1.5mm}{
			\begin{tabular}{l|ccc}
				\hline
				& \textit{wb} & \textit{hs} & \textit{wb}+\textit{hs} \\ \hline
				HOTA$\uparrow$ & 67.2 & 51.9 & \textbf{68.7} \\ \hline
				MOTA$\uparrow$ & 87.6 & 56.2 & \textbf{87.6}  \\ \hline
				IDF1$\uparrow$ & 77.2 & 56.3 & \textbf{77.9} \\ \hline
		\end{tabular}}
	\end{subtable}
\end{table}

\begin{figure}[!t]
	\centering
	\begin{subtable}[t]{0.45\linewidth}
		\centering
		\includegraphics[align=c, width=1\textwidth]{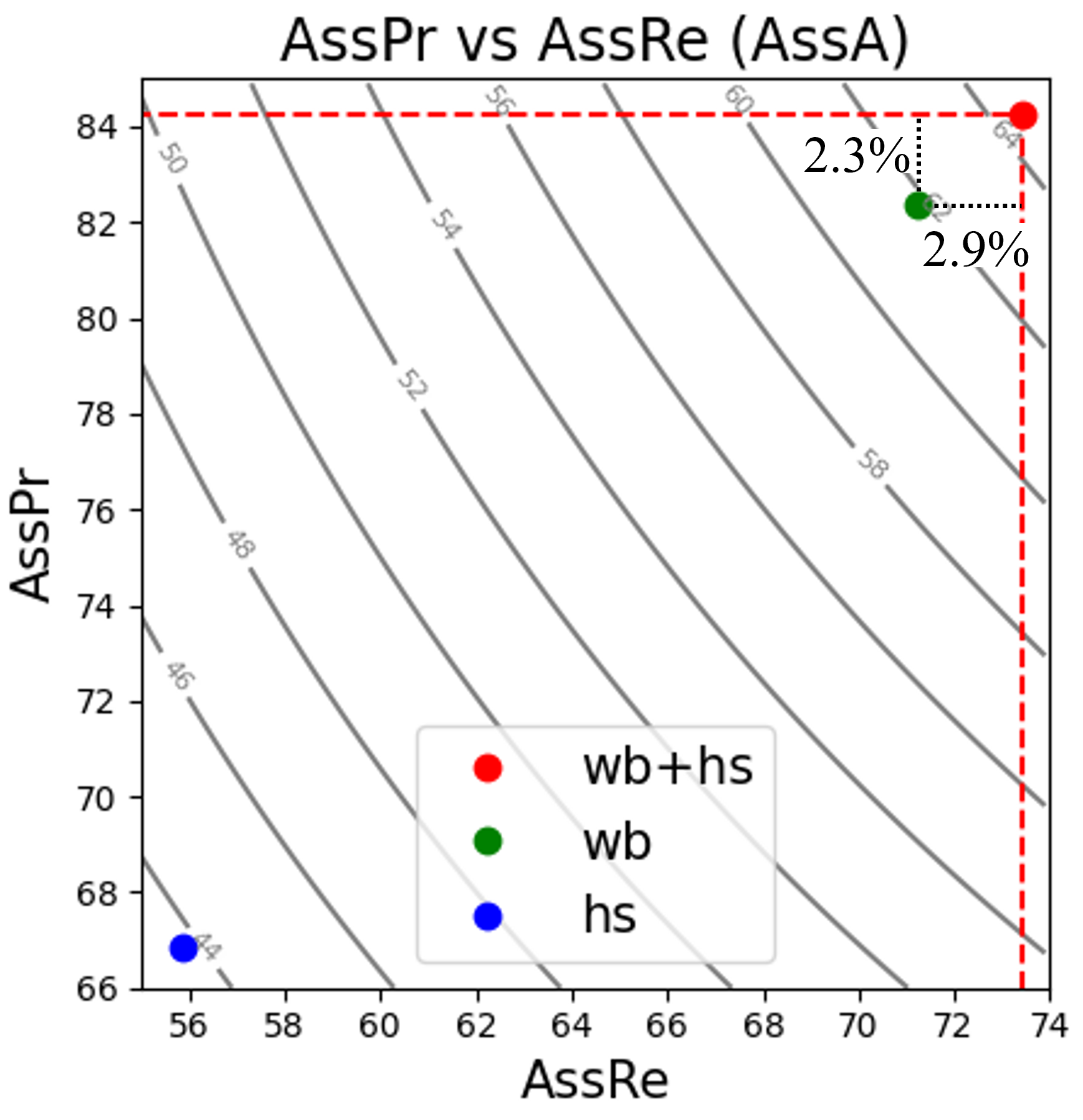}
		\label{assa_tradition}
		\caption{Casual cloth}
	\end{subtable}
	\begin{subtable}[t]{0.45\linewidth}
		\centering
		\includegraphics[align=c, width=1\textwidth]{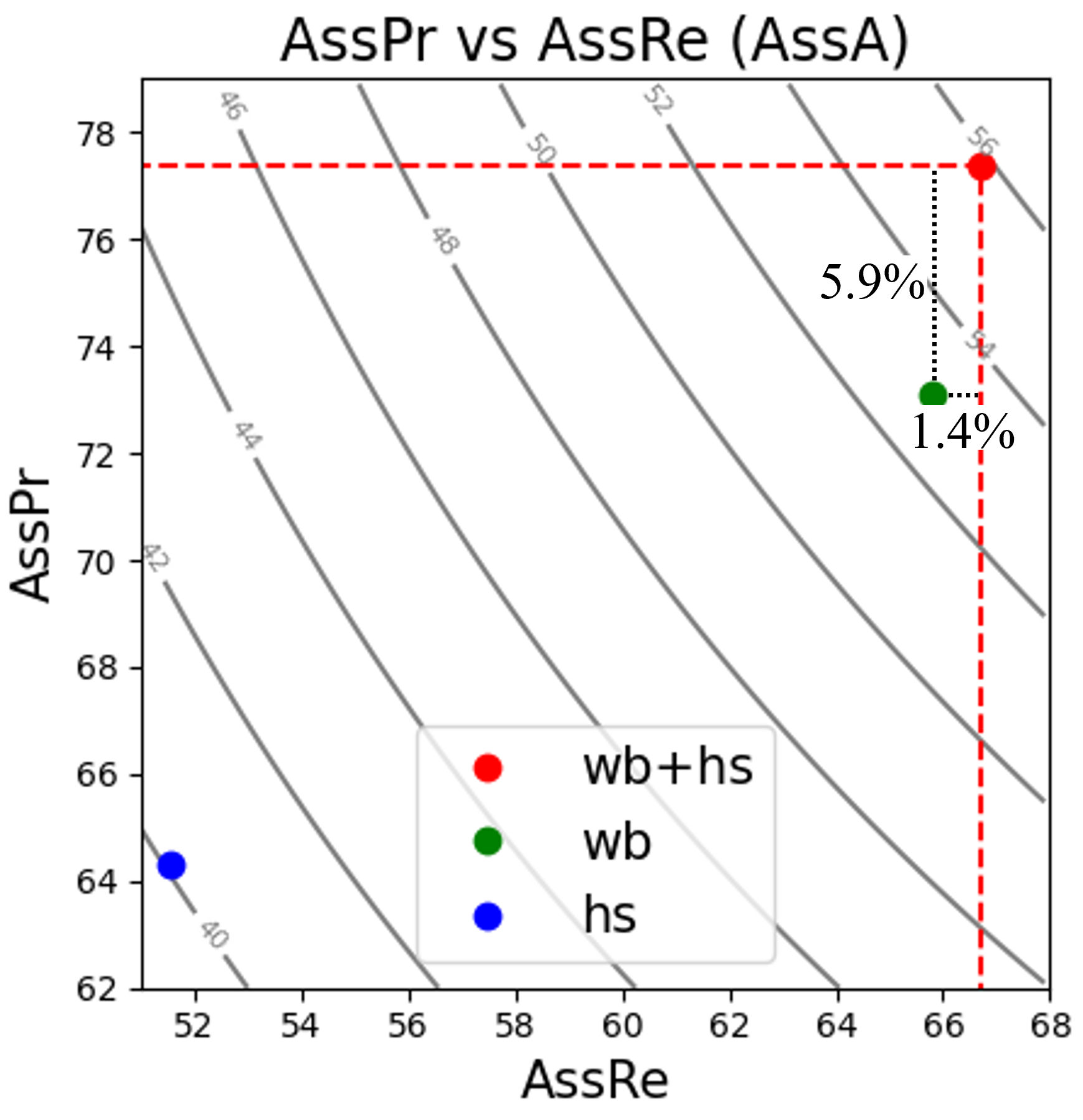}
		\label{assa_similar}
		\caption{Same cloth}
	\end{subtable}
	\caption{Precision and recall scores of data association. The integration of head shoulder contributes equally to precision and recall scores in casual-cloth sub-dataset while contributing more to precision score in same-cloth sub-dataset.}.
	\label{assa}
\end{figure}

To dig deeper into how head shoulder helps improve the overall performance, we show precision and recall scores of association in Fig. \ref{assa}. Compared to the \textit{wb}-only method, for casual-cloth sub-dataset, the percentage promotion of precision and recall after involving head shoulder is 2.3\% and 2.9\%, respectively. Thus, the head shoulder contributes equally to precision and recall. Nevertheless, for same-cloth sub-dataset, the respective promotion is 5.9\% and 1.4\%. Head shoulder contributes much more to precision. In other words, with the help of head shoulder, the tracker does better in identifying different people even if they wear the same. This is due to the discriminative details contained in head shoulder which are untraceable in whole body.

\subsubsection{New Metric}

We show scores of our proposed TGRHOTA in Table \ref{tgrhota}. We can see that TGRHOTA scores are higher than HOTA scores. This is a similar situation as tracker B shown in Fig. \ref{hota_vs_tgrhota}. That is to say, once a target was assigned an ID, our tracker can keep it consistent most of the time.

\begin{table}[th]
	\centering
	\caption{The evaluation result of new TGRHOTA metric.}
	\label{tgrhota}
	\begin{subtable}[t]{0.495\linewidth}
		\caption{Casual Cloth}
		\centering
		\setlength{\tabcolsep}{1.2mm}{
			\begin{tabular}{l|ccc}
				\hline
				& \textit{wb} & \textit{hs} & \textit{wb}+\textit{hs} \\ \hline
				TGRHOTA$\uparrow$ & 74.3 & 55.5 & \textbf{75.2} \\ \hline
		\end{tabular}}
	\end{subtable}
	\begin{subtable}[t]{0.495\linewidth}
		\caption{Same Cloth}
		\centering
		\setlength{\tabcolsep}{1.2mm}{
			\begin{tabular}{l|ccc}
				\hline
				& \textit{wb} & \textit{hs} & \textit{wb}+\textit{hs} \\ \hline
				TGRHOTA$\uparrow$ & 70.1 & 54.9 & \textbf{72.1} \\ \hline
		\end{tabular}}
	\end{subtable}
\end{table}

\subsubsection{FPS}
On a single thread, our method runs at 16fps end-to-end, using an Nvidia 2080Ti GPU card. Parallel running of \textit{wb} and \textit{hs} detectors may speed up the method.

\subsection{Comparison with State-of-the-Art}

We compare our proposal with two newest state-of-the-art methods, CenterTrack \cite{zhou2020tracking} and ByteTrack \cite{zhang2021bytetrack}. Results are shown in Table \ref{sota}. We train CenterTrack and ByteTrack using our dataset and fine-tune some hyper-parameters. Our method achieves the best performance. We show comparison results for reference only, as we have carefully tailored DeepSORT to fit the tour-guide scenario, especially for data association. Designed for video surveillance, it is foreseeable that CenterTrack and ByteTrack perform poorly without any scenario-oriented modifications. 

\begin{table}[!t]
	\centering
	\caption{Comparison with State-of-the-Art.}
	\label{sota}
	\setlength{\tabcolsep}{0.1mm}{
		\begin{tabular}{l|cccc|cccc}
			\hline
			& \multicolumn{4}{c|}{Casual Cloth} & \multicolumn{4}{c}{Same Cloth} \\ \cline{2-9}
			Method & MOTA & IDF1 & HOTA & TGRHOTA & MOTA & IDF1 & HOTA & TGRHOTA \\ \hline
			CenterTrack & 88.1 & 25.1 & 32.0 & 39.3 & 87.7 & 22.0 & 29.8 & 35.5 \\
			ByteTrack & 87.7 & 26.8 & 33.1 & 39.7 & 87.3 & 23.7 & 30.8 & 34.5 \\
			Ours & 87.2 & \textbf{84.2} & \textbf{72.7} & \textbf{75.2} & 87.6 & \textbf{77.9} & \textbf{68.7} & \textbf{72.1} \\ \hline
	\end{tabular}}
\end{table}

\section{Conclusion}

We release the TGRDB dataset, the first large-scale dataset for the applications of TGR. It is captured using a TGR in an indoor tour-guide scenario. We annotate the dataset with whole-body and head-shoulder bounding boxes, as well as unique IDs for each participant. Frequently occlusion, long-term missing and fined-grained targets (dress the same) are the main challenges in this dataset. We believe that TGRDB will help future research in service robotics, long-term multi-person tracking, and find-grained or clothes-inconsistency person re-identification. Along with the dataset, we propose a more practical metric, TGRHOTA, to evaluate trackers in the tour-guide scenario. Different from existing metrics, TGRHOTA only considers TPs at most previous frames, which is more applicable in TGR. As part of our work, we propose TGRMPT, a novel head-shoulder aided multi-person tracking system that leverages best of discriminative cues contained in head shoulder which are untraceable in whole body. Extensive experiments have confirmed the significant advantages of our proposal.

\section*{Acknowledgment}

This work is supported in part by the Cloud Brain Foundation Grants U21A20488 in Zhejiang Lab. The authors would also thank Fellow Academician Jianjun Gu for establishing and leading the Tour-Guide Robot project.

\bibliography{bib.bib}

\end{document}